\documentclass[conference]{IEEEtran}
\IEEEoverridecommandlockouts

\usepackage{graphicx}
\usepackage{amsmath,amssymb}
\usepackage{booktabs}
\usepackage{algorithm}
\usepackage{algorithmic}
\usepackage{xcolor}
\usepackage{url}
\usepackage[hidelinks]{hyperref}
\usepackage{orcidlink}
\usepackage[caption=false]{subfig}
\usepackage{dblfloatfix}

\graphicspath{{./}}

\newcommand{\floor}{$\blacklozenge$}

\begin{document}

\title{Open-World Hierarchical Perception: Taxonomic\\
Abstraction over Class-Agnostic Proposals for the Safe\\
Handling of Out-of-Vocabulary Road Objects}

\author{\IEEEauthorblockN{Felix Schaller \orcidlink{0000-0002-3218-3214}
\href{https://orcid.org/0000-0002-3218-3214}{0000-0002-3218-3214}}
\IEEEauthorblockA{Independent Researcher\\
Dubai, UAE / Munich, Germany\\
Email: \href{mailto:inquiry@felixschaller.com}{inquiry@felixschaller.com}\\
Third paper in a series; v1 archived at Zenodo: \href{https://doi.org/10.5281/zenodo.21593472}{doi:10.5281/zenodo.21593472}}}

\maketitle

\begin{abstract}
A closed-set object detector for autonomous driving must assign every object one
of a fixed set of class labels. On an object outside that set (a horse-drawn
carriage, a piece of road debris, livestock on a rural road) it can only force a
confident but wrong specific label or drop the object. Prior work in this series
replaced the flat label set with a \emph{hierarchical taxonomy} and a runtime
\emph{abstraction} rule, but evaluated it only on the boxes a closed detector
already produces. This paper takes the layer \emph{open-world}. We (i) place the
taxonomic abstraction layer on top of class-agnostic region proposals so that
objects the closed detector never boxes can still be classified or flagged; (ii)
report a feasibility study of three open-world signals (class-agnostic
segmentation, appearance-based out-of-distribution scoring, and monocular
depth) that shows why no single two-dimensional cue is sufficient and how they
compose; and (iii) run the evaluation the earlier papers could not: a
\emph{ground-truth leave-classes-out} benchmark on real annotated objects. Holding
out seven COCO classes from the taxonomy and classifying their $235$ ground-truth
crops, a flat closed head emits a confident wrong specific label $100\%$ of the
time ($37\%$ of them in the wrong super-category, e.g.\ an animal named as a
vehicle), whereas the hierarchical layer emits \emph{zero} confident wrong
specific labels and safely handles $94\%$ of the objects (a correct
super-category, or an explicit \textsc{unknown obstacle}). We are explicit that
this is a \emph{safety} result, not a specificity one: the correct super-category
is recovered only $26\%$ of the time and the remaining $69\%$ are conservatively
flagged unknown. The contribution is an open-world perception layer that never
makes a confident categorical mistake on an out-of-vocabulary object, together
with an honest account of its cost.
\end{abstract}


\section{Introduction}
Perception for automated driving is dominated by detectors trained on a
\emph{closed} vocabulary of object classes. This is efficient and accurate for the
categories in the training set, but it makes the open world invisible: an object
whose class was never in the label set has, from the detector's point of view, no
correct answer. The detector must either force the nearest in-vocabulary label (a
truck named as a car, a fallen tree named as nothing at all) or suppress the
object. In an open-set safety context both failures are categorical: the wrong
size, mass and behaviour model is attached to a real obstacle, or the obstacle is
dropped.

The premise of this series~\cite{schaller2025}, grounded in the observation that
``patterns are everywhere''~\cite{schaller_patterns}, is that perception should be
allowed to be \emph{less specific but still correct} rather than forced to be
specific and wrong. Concretely, the flat label list is replaced by a hierarchical
taxonomy (\dots{}$\to$ Truck $\to$ Transport Vehicle $\to$ Vehicle $\to$ Living
Being / Static Object), and a runtime rule lets a detection descend the taxonomy
only as far as the visual evidence justifies. A per-branch \emph{safety floor}
bounds this fallback so the system never collapses into a useless generic bucket;
below the floor it emits an explicit, localized \textsc{unknown obstacle}. The
first two papers established this on closed-detector boxes (v1) and added a second,
segmentation-based perception path that cross-validates each box (v2).

Both earlier evaluations, however, share a structural limitation that we stated as
an open problem: the ``novel'' objects were a \emph{label-space proxy}. They were
COCO classes we simply left out of our taxonomy (giraffe, zebra), objects a
standard detector \emph{does} recognize. On such data one cannot fairly claim to
handle novelty better than the closed detector, because the closed detector is not
actually failing. Closing that gap requires two things this paper provides: an
open-world \emph{front-end} that produces regions for objects the closed detector
would not box, and a \emph{ground-truth} evaluation on objects that are genuinely
outside the model's vocabulary.

\begin{figure}[t]
\centering
\includegraphics[width=\columnwidth]{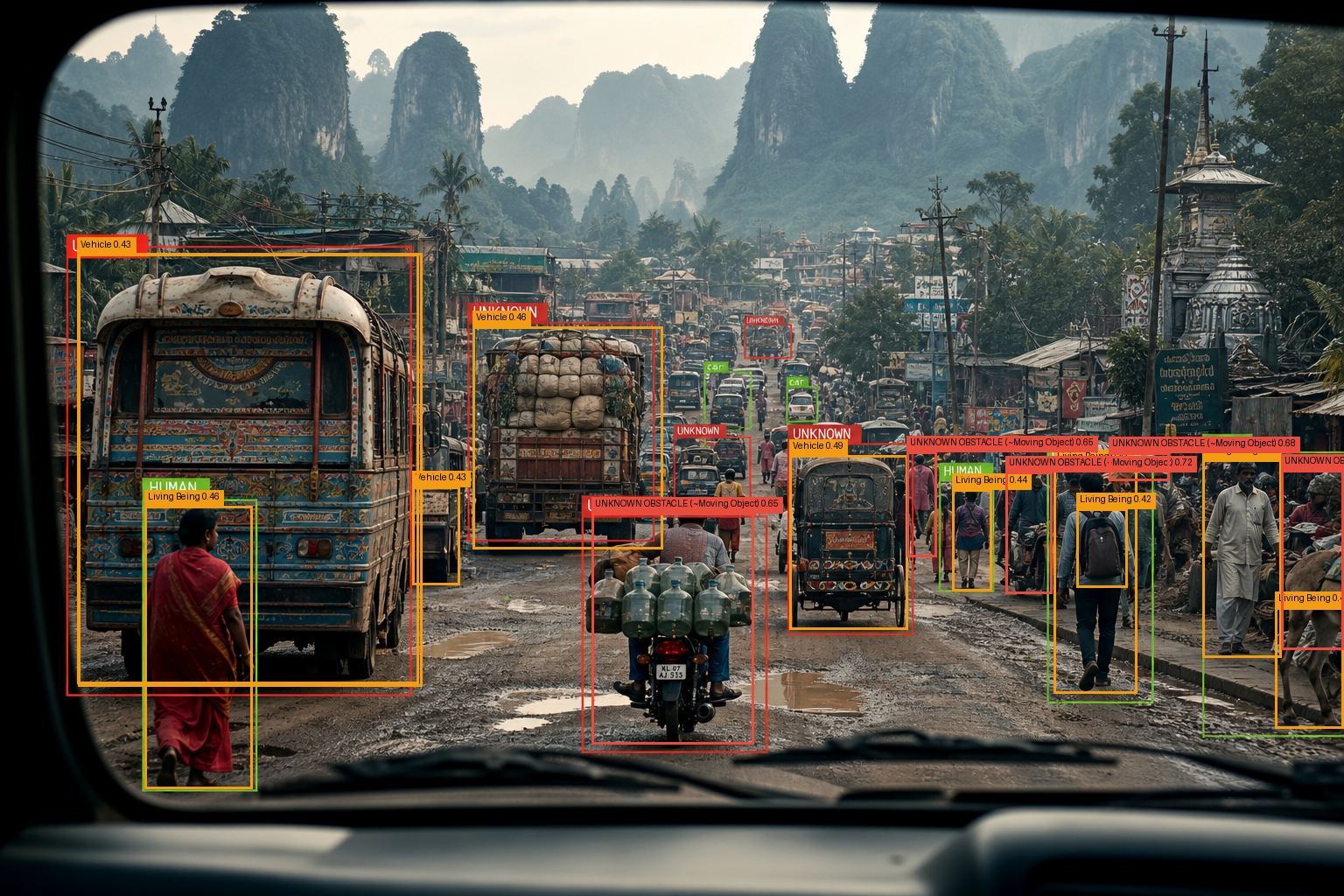}
\caption{The hierarchical layer on a road scene. Each region carries the most
specific \emph{safe} taxonomy level the evidence supports, or an explicit
\textsc{unknown obstacle}; it never attaches a confident wrong specific label.}
\label{fig:title}
\end{figure}

\noindent\textbf{Contributions.}
\begin{itemize}
\item An \emph{open-world} perception layer (\S\ref{sec:method}) that runs the
taxonomic abstraction rule over class-agnostic region proposals, so objects the
closed detector misses can still receive a safe hierarchical label or be flagged.
\item A feasibility study (\S\ref{sec:feasibility}) of three open-world signals
(class-agnostic segmentation, appearance OOD, monocular depth), including two
honest negative results, showing that no single two-dimensional cue suffices and
that recall (segmentation) and precision (geometry) are complementary.
\item A \emph{ground-truth leave-classes-out} benchmark
(\S\ref{sec:benchmark}) on $235$ real out-of-vocabulary objects: the flat head is
confidently wrong $100\%$ of the time ($37\%$ in the wrong super-category); the
hierarchical layer is confidently wrong $0\%$ of the time and safely handles
$94\%$. We report this as a safety result and quantify its cost (a $69\%$
conservative-abstention rate).
\end{itemize}

\section{Related Work}
\textbf{Open-set and open-world recognition.} Open-set recognition~\cite{scheirer2013}
formalizes the requirement that a classifier reject inputs unlike its training
classes rather than force a known label; open-world recognition adds the
incremental discovery of new categories. Our contribution is orthogonal to the
scoring rule: instead of a binary known/unknown decision we place the reject option
at \emph{every level} of a semantic hierarchy, so the system can also answer
``some kind of vehicle'' when it can neither name the leaf nor honestly call the
object unknown.

\textbf{Open-vocabulary and class-agnostic detection.} Open-vocabulary detectors
such as YOLO-World~\cite{cheng2024yoloworld} and Grounding
DINO~\cite{liu2023groundingdino} box objects from a text prompt, and the Segment
Anything family~\cite{kirillov2023}, in particular the lightweight
MobileSAM~\cite{zhang2023mobilesam}, proposes class-agnostic masks for
\emph{everything} in a scene. These give recall on untrained objects but no
semantics and no notion of safe abstraction; we use class-agnostic proposals as
one front-end and supply the semantics and the safety floor on top.

\textbf{Hierarchical classification and depth.} Hierarchy-aware
classifiers~\cite{bertinetto2020} reduce the \emph{severity} of mistakes by making
errors land near the truth in a taxonomy, a property long noted in the psychology
of basic-level categories~\cite{rosch1975}. Zero-shot open-vocabulary
scoring~\cite{radford2021} lets us attach a score to any taxonomy node without
training. Monocular depth estimators~\cite{yang2024depthanything} provide a
per-pixel geometric cue that we test as a precision filter. Autonomous-driving
anomaly benchmarks~\cite{chan2021,li2022} supply corner-case imagery; the COCO
dataset~\cite{lin2014coco} supplies the ground-truth boxes and labels used for our
leave-classes-out evaluation.

\section{Method: Open-World Hierarchical Perception}
\label{sec:method}
The system is a pipeline of three separable stages, \emph{propose},
\emph{classify}, \emph{validate}, so that the open-world extension slots in as new
proposers and validators without disturbing the semantic core.

\subsection{Taxonomy and Safety Floor}
\begin{figure*}[t]
\centering
\includegraphics[width=0.96\textwidth]{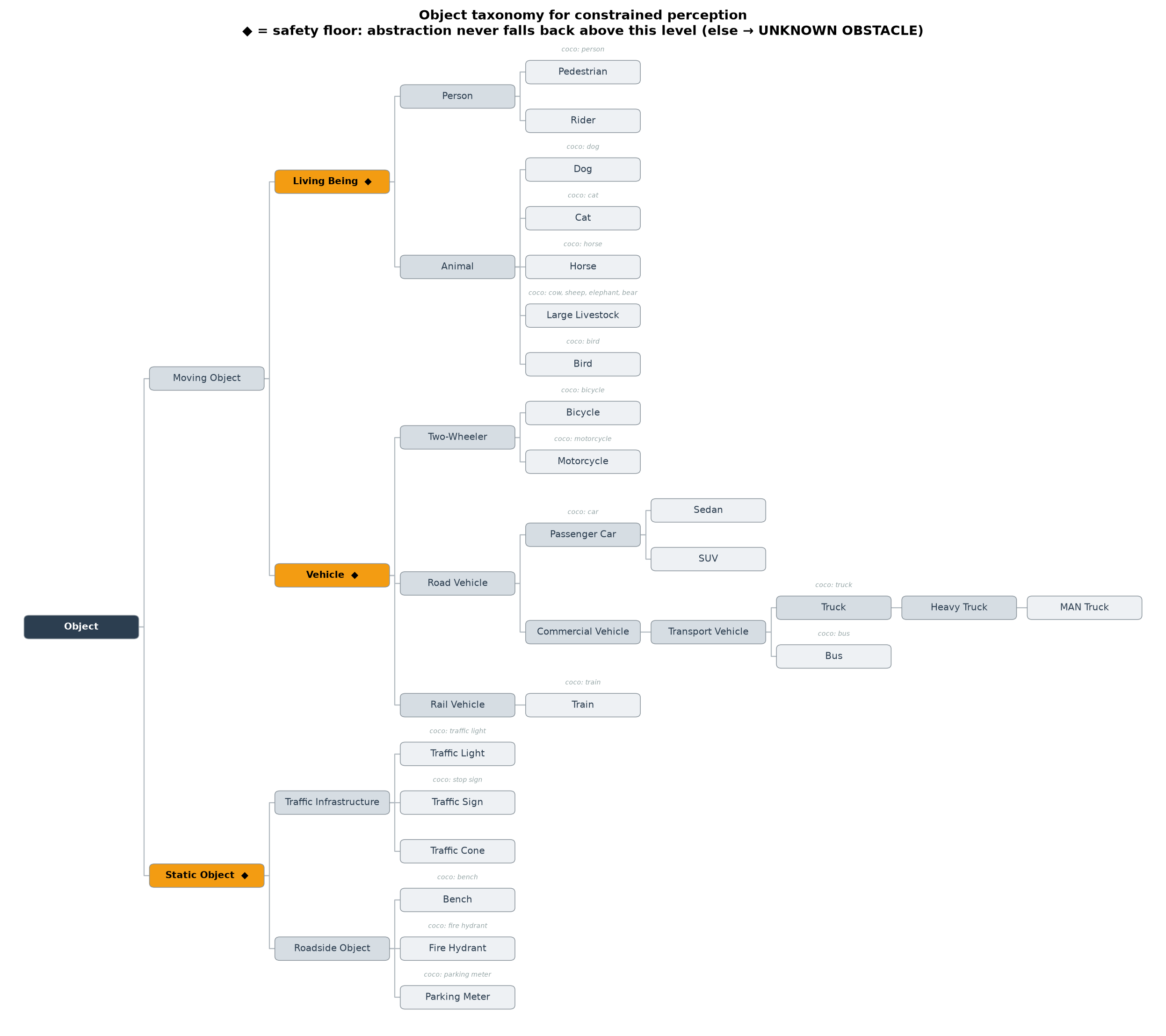}
\caption{The object taxonomy and the per-branch safety floor (\floor). A detection
descends only as far as the evidence justifies; the floor bounds the fallback so
abstraction stays actionable, and below it the object is flagged unknown rather
than labelled with a useless generic category.}
\label{fig:tax}
\end{figure*}

Objects are organized in a directed tree whose leaves are concrete classes (Sedan,
Cyclist, Horse) and whose internal nodes are safety-relevant abstractions (Vehicle,
Living Being, Static Object). Each branch declares a \emph{floor}: the coarsest
level that is still actionable for planning. Abstraction is allowed down to the
floor; anything below it is reported as \textsc{unknown obstacle} rather than as a
too-generic ``object'' (Fig.~\ref{fig:tax}).

\subsection{Proposers: from Closed Boxes to Class-Agnostic Regions}
The closed front-end runs a pretrained YOLO detector~\cite{redmon2016} at high
recall (low confidence, class-agnostic non-maximum suppression) to obtain candidate
boxes. Because a closed detector will not box a genuinely unfamiliar object, the
open-world front-end adds a \emph{class-agnostic} proposer: MobileSAM
\cite{zhang2023mobilesam,kirillov2023} segments regions regardless of category, so
an untrained object still yields a region to classify. The two proposal sources are
complementary and are simply unioned before classification.

\subsection{Zero-Shot Leaf Scoring and Mass-Aggregated Abstraction}
For each region we take the CLIP~\cite{radford2021} image embedding and its cosine
similarity to a text prompt for every taxonomy leaf, giving a softmax distribution
over leaves at temperature $\tau$. Rather than take the arg-max leaf, we
\emph{aggregate} leaf mass upward: the score of an internal node is the sum of the
mass of its descendant leaves. Starting at the root, we descend to a child only
while that child concentrates at least a commit fraction $m$ of the local mass;
where the mass splits, we stop and report the current node, a justified
abstraction. If even the best leaf's absolute similarity is below a floor
$s_{\min}$, or descent halts below the branch safety floor, the region is reported
as \textsc{unknown obstacle}. This is the mechanism that turns an uncertain
distribution into a \emph{coarser but correct} label instead of a confident wrong
leaf (parameters in \S\ref{sec:benchmark}).

\subsection{Geometry as a Precision Filter}
Class-agnostic proposals over-generate (sky, road, vegetation, object parts). A
monocular depth map~\cite{yang2024depthanything} provides an appearance-independent
cue, foreground-vs-background separation and a flat-vs-solid test, to suppress
background regions and, in principle, flat ``billboard'' fakes. We treat geometry
as an optional precision filter on the proposal stream; \S\ref{sec:feasibility}
reports how far a purely two-dimensional version of this signal actually gets.

\section{Open-World Feasibility Study}
\label{sec:feasibility}
Before committing to an architecture we measured each open-world signal in
isolation. The results are deliberately reported with their negatives, because the
negatives determine the design.

\subsection{Class-Agnostic Proposals: Recall Up, Precision Down}
Feeding MobileSAM regions into the hierarchical classifier on anomaly
imagery~\cite{chan2021}, the closed YOLO front-end produced $19$ detections on a
sample where MobileSAM added $148$ further regions the detector had missed. The
hierarchical \textsc{unknown} gate correctly filtered $112$ of these as unknown, its
value as a filter, but the $36$ that received a category were mostly background
(vegetation labelled ``Living Being''; Fig.~\ref{fig:sam}). Class-agnostic
segmentation therefore delivers open-world \emph{recall} but, on its own, poor
precision.

\subsection{Appearance-Based OOD: a Negative Result}
We tested whether an appearance-only out-of-distribution score, the margin between
the best taxonomy-leaf similarity and the best ``background / none-of-these'' prompt
similarity, could separate implausible detections from plausible ones on the Road
Anomaly set. It does not: the margins of implausible, plausible-out-of-taxonomy and
good in-taxonomy cases overlap almost entirely, and the best threshold that catches
all implausible cases also breaks a third of the good ones. Tellingly, a full-frame
aircraft matched ``a plain textureless background'' more strongly than ``an
aircraft'', CLIP on the 2D crop cannot reliably tell what the object is in the
first place. The ambiguity is not lexical; it is one of scale and undersampling.

\subsection{Monocular Depth: Real Signal, Scale-Limited}
\begin{figure}[t]
\centering
\subfloat[Class-agnostic proposals: high recall, but many background regions
survive with a category (here vegetation as ``Living Being'').]{%
\includegraphics[width=\columnwidth]{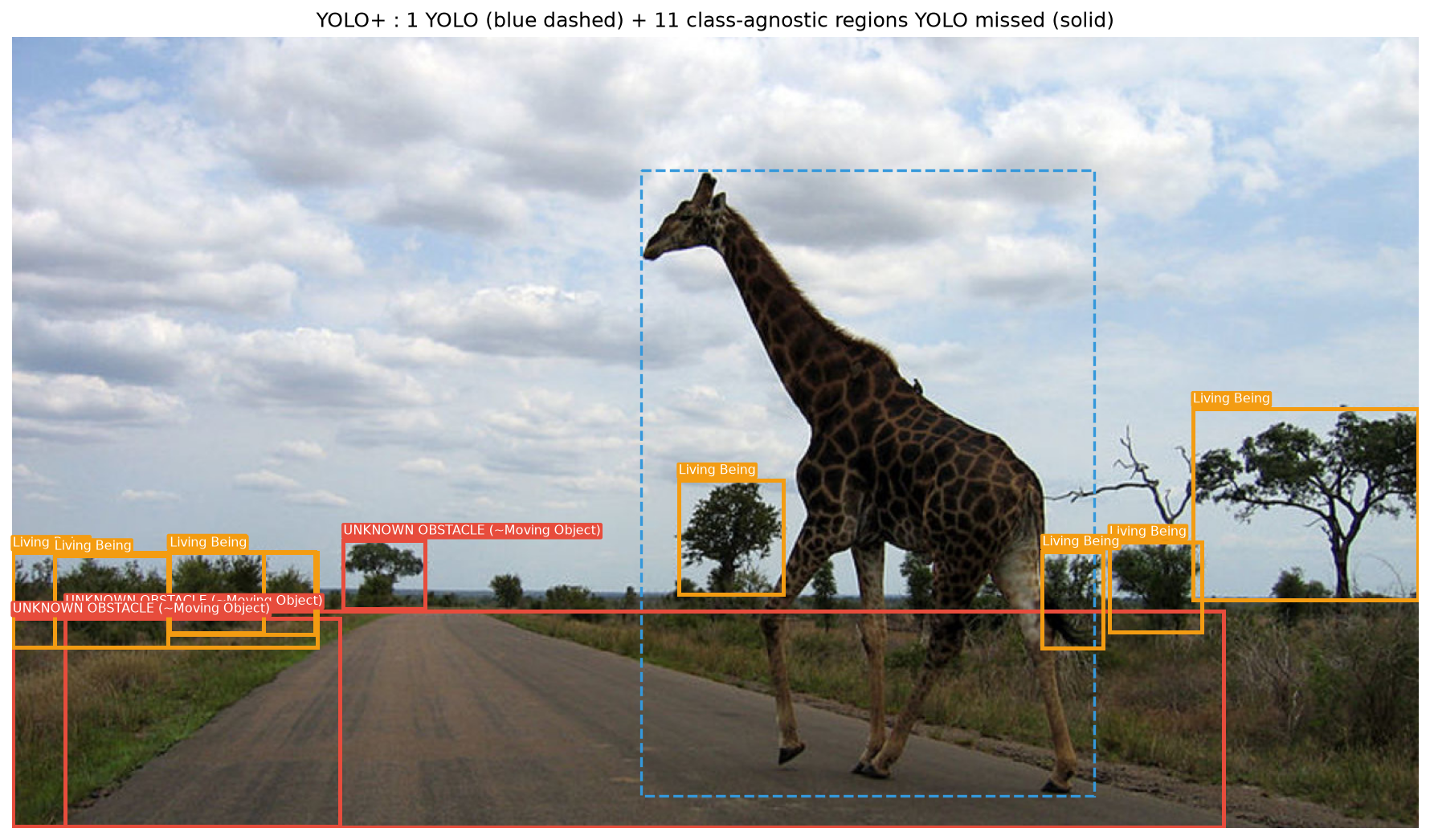}}\\
\subfloat[Monocular depth separates foreground from background, an
appearance-independent precision cue, but only above a scale.]{%
\includegraphics[width=\columnwidth]{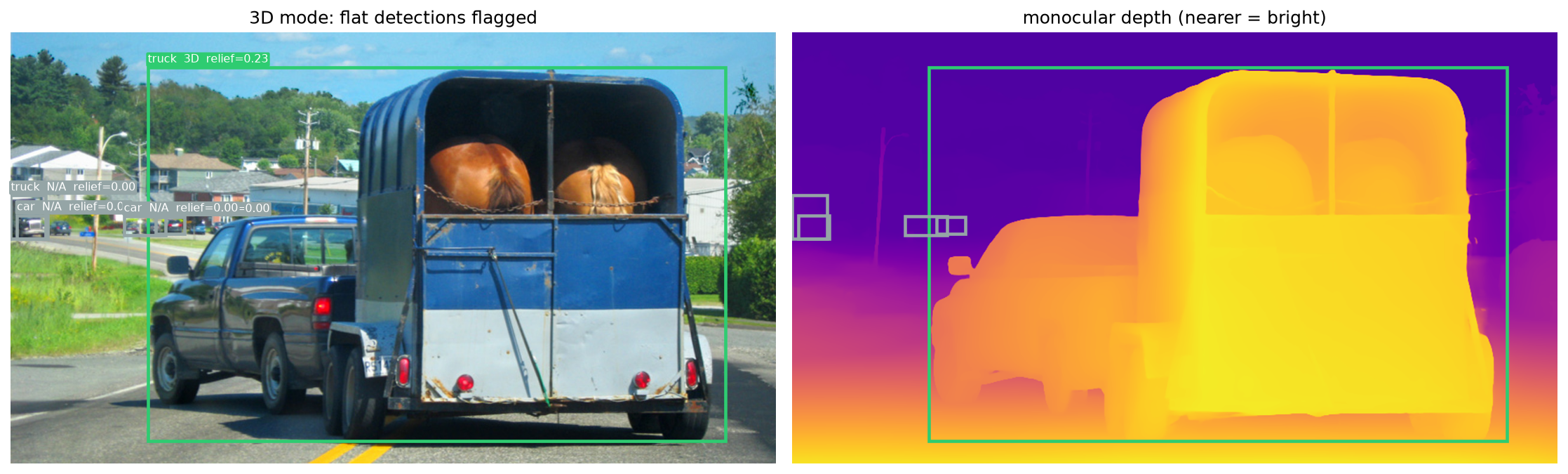}}
\caption{Open-world signals in isolation (\S\ref{sec:feasibility}): recall from
segmentation (a) and a geometric precision cue from depth (b).}
\label{fig:sam}
\label{fig:depth}
\end{figure}

A per-region flatness/foreground test on a monocular depth map cleanly separated
foreground from background (Fig.~\ref{fig:depth}) and, behind an assessability gate,
read large upright objects as three-dimensional while holding small or distant
regions as ``not assessable'' rather than misflagging them. But relief is
shape-sensitive (a compact animal read as flat), and monocular depth is relative,
not metric. The recurring confound is again \emph{scale}: geometric structure is
only measurable above a certain apparent size, which is exactly where metric LiDAR
would help.

\subsection{Synthesis}
The three signals are complementary rather than competing: class-agnostic
segmentation supplies recall, geometry is the natural precision filter for it, and
appearance OOD does not stand alone. A dependable open-world front-end therefore
needs proposals \emph{and} a geometric filter; the hierarchical abstraction supplies
the semantics and the \textsc{unknown} safety net on top. The abstraction layer is
already dependable, which is why the evaluation that follows isolates it.

\section{Leave-Classes-Out Evaluation}
\label{sec:benchmark}
The evaluation the earlier papers could not run needs objects that are genuinely
out-of-vocabulary \emph{and} carry ground-truth labels, so that ``correct
super-category'' is measured rather than asserted.

\subsection{Protocol}
We use the COCO validation set~\cite{lin2014coco}, which provides ground-truth boxes
and class labels. We designate seven classes as held-out and \emph{remove their
leaves from the taxonomy}: \{truck, bus\} (true super-category Vehicle) and \{horse,
cow, sheep, elephant, bear\} (true super-category Living Being). These classes are
now out-of-vocabulary, but because they are ground-truth we still know the correct
super-category. We stream $800$ validation images and, for every ground-truth object
of a held-out class larger than $1\%$ of the frame, crop the ground-truth box and
classify it two ways on the \emph{pruned} taxonomy:
\begin{itemize}
\item \textbf{Flat / closed head:} the arg-max leaf of the CLIP distribution, the
behaviour of a closed classifier that must name a specific class.
\item \textbf{HOWC:} the mass-aggregated abstraction of \S\ref{sec:method}
($m=0.40$, $\tau=0.06$, $s_{\min}=0.20$), which may return a leaf, an internal
super-category, or \textsc{unknown}.
\end{itemize}
This isolates the \emph{classification} half of the open-world claim: a good
detector boxes these objects, and the question is what a fixed vocabulary can
\emph{say} about them. The procedure yields $n=235$ out-of-vocabulary ground-truth
objects.

\subsection{Metrics and Results}
\begin{figure}[t]
\centering
\includegraphics[width=\columnwidth]{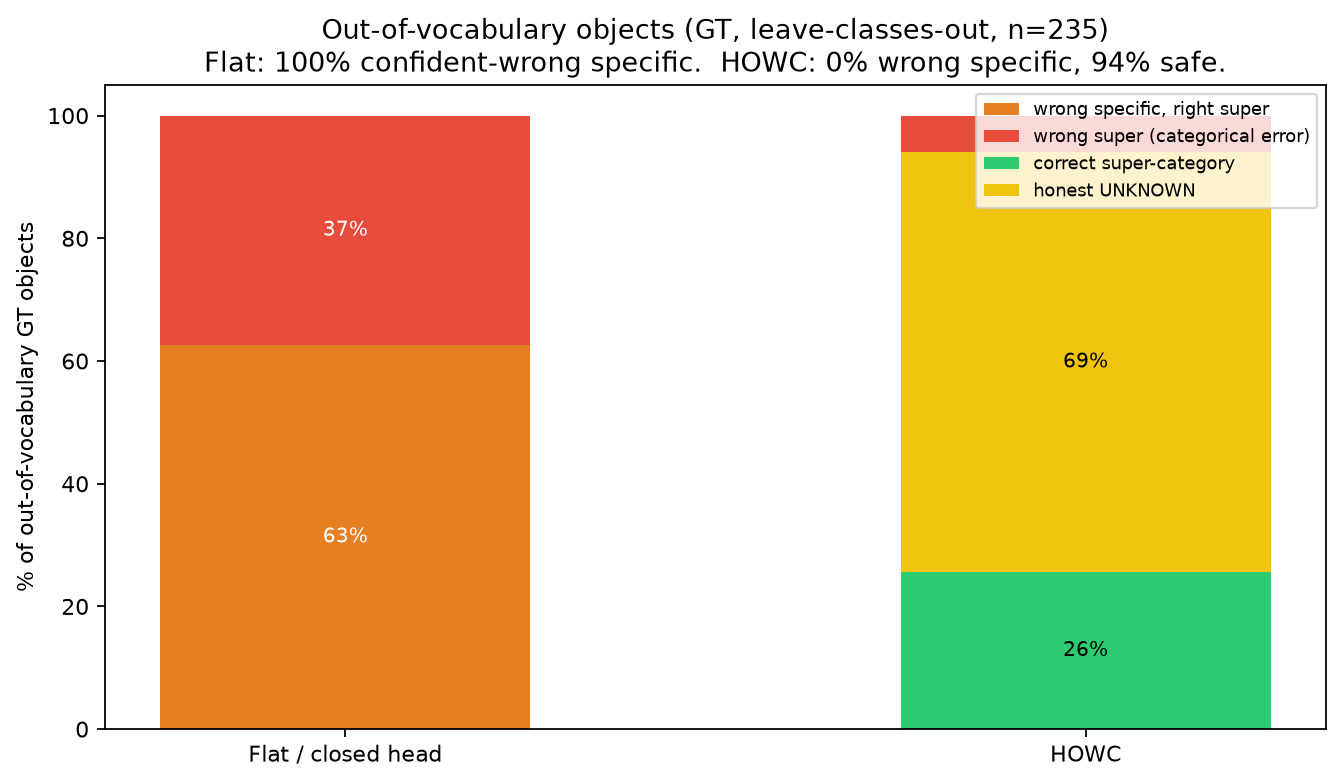}
\caption{Outcome distribution on out-of-vocabulary ground-truth objects. The flat
head is confidently wrong on every object ($37\%$ of the time in the wrong
super-category); HOWC makes no confident wrong specific claim and is safe on $94\%$,
though most of that safety is conservative abstention ($69\%$ unknown) rather than a
recovered super-category ($26\%$).}
\label{fig:owb}
\end{figure}

For each object we know the true super-category, so we can classify every outcome as
safe or unsafe. A flat prediction is scored by whether its leaf lies on the correct
branch (still a wrong specific label, but at least the right super-category) or on
the wrong branch (a categorical error). A HOWC prediction is \emph{safe} if it is
the correct super-category or an honest \textsc{unknown}, and \emph{unsafe} if it
over-commits to a wrong leaf or abstracts to the wrong super-branch.
Table~\ref{tab:owb} and Fig.~\ref{fig:owb} report the outcome distribution.

\begin{table}[t]
\caption{Outcomes on $n=235$ out-of-vocabulary ground-truth objects
(leave-classes-out on COCO val).}
\label{tab:owb}
\centering
\begin{tabular}{lcc}
\toprule
Outcome & Flat / closed head & HOWC \\
\midrule
Confident wrong specific label & $100\%$ & $\mathbf{0\%}$ \\
\quad of which wrong super-branch & $37\%$ & n/a \\
Correct super-category & n/a & $26\%$ \\
Honest \textsc{unknown} & n/a & $69\%$ \\
Wrong super-branch & n/a & $6\%$ \\
\midrule
\textbf{Safely handled} & $\mathbf{0\%}$ & $\mathbf{94\%}$ \\
\bottomrule
\end{tabular}
\end{table}

\subsection{Honest Reading}
The headline is a \emph{safety} result, and we are careful not to overstate it. HOWC
never emits a confident wrong specific label ($0\%$ vs.\ the flat head's $100\%$),
and it avoids the flat head's categorical errors, $37\%$ of the flat head's labels
put the object in the wrong super-category, for example an animal named as a
vehicle, against $6\%$ for HOWC. On $94\%$ of out-of-vocabulary objects HOWC is
therefore safe. But safety here is dominated by \emph{abstention}: the correct
super-category is recovered only $26\%$ of the time, while $69\%$ of objects are
conservatively flagged \textsc{unknown}. This is the safety floor behaving as
designed, a flagged unknown obstacle is preferable to a confident wrong guess, but
it is not a claim of superior recognition \emph{accuracy}. The layer's value is that
it converts $100\%$ confident-wrong into $0\%$ confident-wrong at the price of
frequent honest uncertainty, which in a safety context is the trade one wants.

\subsection{Corroboration on In-Vocabulary Objects}
For completeness we confirm the same mechanism does no harm when the object \emph{is}
in vocabulary. On in-taxonomy anomaly-set objects the mass-aggregated layer produces
$0\%$ off-branch (categorical) errors with $24\%$ calibrated abstention, versus a
flat arg-max head's $53\%$ off-branch errors on the same boxes, i.e.\ it trades some
specificity for the elimination of categorical mistakes, not for a loss of correct
answers.

\section{Discussion}
\textbf{Safety, not specificity.} The consistent finding across both the
out-of-vocabulary and in-vocabulary settings is that hierarchical abstraction buys
the elimination of \emph{confident categorical mistakes}, and pays for it in
specificity and abstention. For automated driving that is the right side of the
trade: a planner can act on ``some large living being ahead'' or on ``unknown
obstacle, localized here'', but a confident ``sedan'' attached to a standing horse
is a silent, unrecoverable error.

\textbf{The cost of conservatism.} The $69\%$ abstention rate is high and is the
principal limitation of the current system. Its cause is the same scale/undersampling
effect the feasibility study exposed: on many out-of-vocabulary crops the CLIP mass
does not concentrate enough on any branch to justify committing even to a
super-category, so the layer defaults to \textsc{unknown}. Sharpening this, so that
more objects earn a correct coarse category and fewer fall through to
unknown, without reintroducing confident errors is the central open problem, and is
exactly where metric geometry (\S\ref{sec:feasibility}) and motion cues are expected
to help.

\textbf{A hierarchical data engine.} Because the layer emits fine labels where it is
confident, coarse labels where only abstraction holds, and \textsc{unknown} where it
is not, running it over unlabelled driving video produces a \emph{hierarchically}
labelled corpus with a built-in review queue (the unknowns). That corpus is the
natural training input for a future detector trained directly on the taxonomy,
closing the loop from open-world handling back to closed-set accuracy.

\textbf{Limitations.} The evaluation isolates classification given ground-truth
boxes; an end-to-end open-world system also depends on the recall/precision of the
proposal front-end, which \S\ref{sec:feasibility} shows is not yet dependable in
pure 2D. Monocular depth is relative, not metric. The held-out set is seven COCO
classes; broader taxonomies and true corner-case objects remain to be tested. And
the abstention rate must come down for the coarse labels to be useful, not merely
safe.

\textbf{Naming.} ``HOWC'' (Hierarchic Open-World Classifier) is an internal working
name. YOLO is a trademark; any public release will use a distinct name to avoid
confusion.

\section{Conclusion and Future Work}
We took a hierarchical taxonomic abstraction layer from classifying a closed
detector's boxes to open-world perception, and evaluated it on the axis the earlier
papers could not: real, ground-truth, out-of-vocabulary objects. Against a flat
closed head that is confidently wrong on $100\%$ of such objects ($37\%$ of them
categorically), the layer makes \emph{no} confident wrong specific claim and handles
$94\%$ safely, while paying for that safety with a $69\%$ honest-abstention rate
that we report rather than hide. Future work is directed at that rate: metric
geometry from LiDAR or stereo, and motion cues, to convert conservative unknowns
into correct coarse categories without reintroducing categorical mistakes; a
dependable proposal front-end that pairs class-agnostic recall with a geometric
precision filter; and the hierarchical data engine that turns the layer's own
open-world output into training data for a taxonomy-native detector.

\section*{Reproducibility}
Code, taxonomy, the leave-classes-out benchmark script, and the full history are
available at \url{https://github.com/freshNfunky/IE2025-Research-Paper}. The
benchmark of \S\ref{sec:benchmark} is \texttt{scripts/v3\_openworld\_benchmark.py}.

\bibliographystyle{IEEEtran}
\bibliography{bib,bib_v3}

\end{document}